\documentclass[accepted, mathfont=cm]{uai2021} 

\usepackage[american]{babel}

\usepackage{natbib} 
\usepackage{mathtools} 
\usepackage{booktabs} 
\usepackage{tikz} 

\usepackage[utf8]{inputenc} 
\usepackage[T1]{fontenc}    
\usepackage{hyperref}       
\usepackage{url}            
\usepackage{booktabs}       
\usepackage{amsfonts}       
\usepackage{nicefrac}       
\usepackage{microtype}      

\usepackage{mathtools}      
\usepackage{amsmath}        
\usepackage{amsthm}         
\newtheorem{Theorem.}{Theorem}
\newtheorem{Definition.}{Definition}
\newtheorem{Proposition.}{Proposition}
\usepackage{amssymb}        
\usepackage[ruled,vlined,linesnumbered]{algorithm2e}
\usepackage{braket}                               
\newcommand{\argmin}{\mathop\textrm{argmin}\limits} 

\usepackage{thmtools}
\declaretheoremstyle[
  spaceabove=0pt, spacebelow=0pt,
  headfont=\itshape,
  notefont=,
  notebraces={(}{)},
  postheadspace=1em,
  numbered=no,
  qed=$\square$
]{myproof}
\declaretheorem[title=Proof, style=myproof]{myproof}
\renewenvironment{proof}{\begin{myproof}}{\end{myproof}}

\renewcommand{\cite}{\citep}

\newcommand\hidemath{r}
\allowdisplaybreaks[1] 
\renewenvironment{itemize}{%
\begin{list}{\textbullet\ \ }{%
    \setlength{\itemindent}{0pt}
    \setlength{\leftmargin}{2em}%
    \setlength{\rightmargin}{0pt}%
    \setlength{\labelsep}{0pt}%
    \setlength{\labelwidth}{3em}%
    \setlength{\itemsep}{0em}%
    \setlength{\parsep}{0em}%
    \setlength{\listparindent}{0pt}%
    \setlength{\topsep}{0em}
}}{\end{list}}
\newcounter{enum2}

\title{Fast Rank Reduction for Non-negative Matrices via Mean Field Theory}

\author[1,2]{{Kazu Ghalamkari}{}} 
\author[1,2,3]{Mahito Sugiyama}
\affil[1]{%
    National Institute of Informatics \\
    Tokyo, Japan
}
\affil[2]{%
    The Graduate University for Advanced Studies, SOKENDAI
}
\affil[3]{JST, PRESTO}

\begin{document}
\maketitle

\begin{abstract}
   We propose an efficient matrix rank reduction method for non-negative 
   matrices, whose time complexity is quadratic in the number of rows or 
   columns of a matrix.
   Our key insight is to formulate rank reduction as a 
   \emph{mean-field approximation} by modeling matrices via a \emph{log-linear model}
   on structured sample space, which allows us to solve the rank reduction 
   as convex optimization. The highlight of this formulation is that the optimal solution that minimizes the KL divergence from a given matrix
   can be \emph{analytically computed in a closed form}. We empirically show that
   our rank reduction method is faster than NMF and its popular variant, 
   lraNMF, while achieving competitive low rank 
   approximation error on synthetic and real-world datasets.
\end{abstract}

\section{Introduction}\label{sec:intro}

\emph{Rank reduction} is a principal technique for matrices to efficiently store and treat them while limiting the loss of information. As we increase the amount of data to be handled, low rank approximation techniques become more and more important and are used in many applications including image
processing~\citep{friedlandFastLowRank2011}, recommender systems~\citep{83071134aff2444e9eb6e81716665924}, and seismic noise attenuation~\citep{chen2017empirical}. Moreover, rank reduction not only reduces the amount of 
information of matrices, but accelerates the subsequent analysis on matrices~\citep{NIPS2000_1866}. For example, \citet{altschulerMassivelyScalableSinkhorn2019} showed that matrix balancing, which is used to compute the entropy regularized optimal transport~\citep{cuturiSinkhornDistancesLightspeed2013}, can be efficiently performed on a rank reduced matrix.


To date, non-negative rank reduction by non-negative matrix factorization (NMF)~\citep{paatero1994positive} has been widely studied since it has various application domains such as audio blind source separation~\cite{leplat2020blind}, video summarization~\cite{liu2019interactive}, and gene expression analysis~\cite{esposito2020nmf}. Variations of NMF have been also studied. For example, lraNMF~\citep{zhou2012fast} is developed to accelerate NMF, and KL-NMF minimizes the KL divergence~\citep{dhillon2005generalized,ho2008non} instead of the least square error that the standard NMF minimizes in approximation of a given matrix.

However, despite its widespread use, its theoretical connection to statistical models remains unrevealed, which is fundamental for further development of non-negative rank reduction techniques. We tackle this problem using \emph{information geometry}~\cite{Amari16}, which enables us to treat different problems across linear algebra and machine learning in a differential geometric manner. In particular, we show that the rank reduction is efficiently achieved as a convex optimization of the \emph{log-linear model}~\cite{Agresti12}, and its special case, rank-$1$ reduction, is understood to be a \emph{mean-field approximation}~\cite{weissHypotheseChampMoleculaire1907}. As a result, we can analytically compute the global optimal rank-$1$ solution which minimizes the KL divergence from an input positive matrix in a closed form. Its formulation leads to a novel rank reduction algorithm, called \emph{Legendre rank reduction}.

To geometrically analyze rank reduction on matrices, we use the log-linear model on a \emph{partially ordered set} (poset)~\citep{Sugiyama16ISIT}. This model has been originally introduced to treat higher-order interactions between variables, and the tight connection between the optimization of the log-linear model and operations on matrices has been demonstrated~\citep{Sugiyama17ICML,Sugiyama2018NeurIPS}. Based on this connection, we formulate rank reduction as a projection in a dually flat statistical manifold, which is equipped with a dual coordinate system $(\theta, \eta)$ connected via Legendre transformation~\citep{Amari16}. 

Moreover, we extend our geometric analysis to the \emph{mean-field approximation}, which has been originally introduced to discuss the phase transitions in ferromagnets~\cite{weissHypotheseChampMoleculaire1907}. The technique of mean-field approximation made an impact across many areas despite the simple way of reducing the many-body problem to the unity problem. It has been widely used in not only physics but also statistics~\cite{petersonMeanFieldTheory}, information theory~\cite{bhattacharyya2000information}, and even in game theory~\cite{cainesLargePopulationStochastic2006,lionsLargeInvestorTrading2007}. In the context of machine learning, mean-field approximation can largely simplify the computation of expected values of Boltzmann machines~\cite{ackley1985learning} avoiding the combinatorial explosion of the computational cost~\cite{anderson1987mean}. \citet{tanakaTheoryMeanField1999} has discussed the mean-field theory in information geometry and pointed out that computation of the expected values from model parameters can be solved as a projection onto a special submanifold, where the transformation between dual parameters $(\theta, \eta)$ is possible in constant time. By capturing the geometric structure of mean-field theory, we present an  unexpected theoretical relationship between rank reduction and mean-field approximation
from the viewpoints of (1) the independence of distributions, and (2) the independence of expected values. 

To summarize, our contribution is three fold:
\begin{itemize}
    \item We propose an efficient non-negative low rank approximation method, called Legendre rank reduction. Its time complexity is quadratic in the number of rows or columns of a matrix.
    \item Our method is formulated as a convex optimization, hence it always finds the globally optimal solution of a KL divergence minimization problem.
    \item By interpreting rank-$1$ reduction as a mean-field approximation, we derive an analytical closed formula to compute the best rank-$1$ approximation for positive matrices in the sense of the KL divergence.
\end{itemize}

\section{The Rank Reduction Algorithm}

We introduce our rank reduction algorithm, called \emph{Legendre rank reduction}. The theoretical aspect of the algorithm will be discussed in Section~\ref{sec:4} while here we focus on the algorithmic aspect of Legendre rank reduction. Throughout the rest of the paper, for an $n \times m$ matrix $\mathbf{A} \in \mathbb{R}^{n \times m}$, we denote the submatrix consisting of all columns from the $a$-th to the $b$-th by $\mathbf{A}_{a:b} \in \mathbb{R}^{n \times (b-a+1)}$.
We assume $m \leq n$ without loss of generality. In the case of $n>m$, our theoretical analysis remains valid by replacing $\mathbf{A}$ with its transpose.

\subsection{Legendre Rank-1 Reduction}\label{sec:21}

Let $\mathbf{A} \in \mathbb{R}^{n\times m}$ be a matrix. We define each entry of Legendre rank-$1$ reduction (L1RR) $\mathrm{L}_1(\mathbf{A}) \in \mathbb{R}^{n\times m}$ of $\mathbf{A}$ as
 \begin{align}\label{eq:l1rr_sol}
    {\rm L}_1(\mathbf{A})_{ij} = 
    \frac{ \left( \sum_{j'} \mathbf{A}_{ij'} \right) \Bigl( \sum_{i'} \mathbf{A}_{i'j} \Bigl) }{ \sum_{i'j'} \mathbf{A}_{i'j'} }.
\end{align}
The time complexity of Legendre rank-$1$ reduction is $O(nm)$. We will show in Section~\ref{sec:4} that the rank of the resulting matrix ${\rm L}_1(\mathbf{A}) \in \mathbb{R}^{n\times m}$ is always $1$ and that the row and the column sum of the matrix do not change during the Legendre rank-$1$ reduction. Interestingly, despite the simple operation of Legendre rank-$1$ reduction, it always gives the best rank-$1$ matrix in the sense of the KL divergence if all elements of $\mathbf{A}$ are strictly larger than 0. That is, we have
\begin{align*}
    \textrm{L}_1(\mathbf{A}) = \text{argmin}_{\mathbf{A}', \mathrm{rank}(\mathbf{A'})=1} D_{KL}( \mathbf{A} ; \mathbf{A'} )
\end{align*}
for any positive matrix $\mathbf{A}$, where the KL divergence is given as $ D_{KL}( \mathbf{A} ; \mathbf{B} ) = \sum_{ij} a_{ij} \log(a_{ij} / b_{ij})$ with $a_{ij} = \mathbf{A}_{ij} / \sum_{ij} \mathbf{A}_{ij}$ and $b_{ij} = \mathbf{B}_{ij} / \sum_{ij} \mathbf{B}_{ij}$. We will give its proof in Theorem \ref{th:L1RR} in Section~\ref{sec:4}. 

The output rank-$1$ matrix ${\rm L}_1(\mathbf{A})$ is in the form of the product of the $n$-dimensional vector whose elements are the row sums and the $m$-dimensional vector whose elements are the column sums of the input matrix $\mathbf{A}$. It takes $nm$ memory to hold the input matrix $\mathbf{A}$ as it is, but after the approximation, the memory requirement is reduced from $nm$ to $n+m$ as we need to hold just these two vectors.

\subsection{Legendre Rank-\protect\hidemath\ Reduction}\label{sec:al}

Legendre rank-$r$ reduction (LrRR) transforms a given real matrix $\mathbf{A} \in \mathbb{R}^{n\times m}$ into a rank reduced matrix $\mathrm{L}_r(\mathbf{A}) \in \mathbb{R}^{n\times m}$. It is archived by applying Legendre rank-$1$ reduction to submatrices of $\mathbf{A}$ according to the following three steps.

\textbf{STEP1} Select $ m - r $ columns and let $C$ be the set of indices of the selected columns $C = \set{c_1,c_2,\dots,c_{m-r}} \subseteq \{1, 2, \dots, m\}$, where we assume $c_i \leq c_{i+1}$. Any columns are valid to obtain a rank reduced matrix. We randomly sample $m - r$ columns without replacement. 

\textbf{STEP2} Partition $C$ into contiguous regions $S_1$, $S_2$, $\dots$, $S_k$.
For example, when $C = \{3,4,5,9,14,15\}$, the partition is obtained as $S_1 = \{3,4,5\}$, $S_2 = \{9\}$, and $S_3 = \{14, 15\}$.

\textbf{STEP3} Replace each submatrix of $\mathbf{A}$ specified by $S_j$ with its L$1$RR. More specifically, for each submatrix $\mathbf{A}_{(\min S_j - 1):\max S_j}$, replace it with $\mathrm{L}_1(\mathbf{A}_{(\min S_j - 1):\max S_j})$. 

The time complexity of LrRR is $O(n(m - r))$. When the target rank $r$ is 1, LrRR agrees with L1RR.

We show that the rank of the resulting matrix $\mathrm{L}_r(\mathbf{A})$ obtained by the above three steps is always smaller than or equal to $r$.
The result of LrRR depends on the selection of columns in \textbf{STEP1}, and we theoretically guarantee that $\mathrm{L}_r(\mathbf{A})$ is the best rank-$r$ approximation \emph{with respect to the selected columns} when $\mathbf{A}$ is a positive matrix.
There may still exist some rank-$r$ matrix $\mathbf{B}$ such that $D_{KL}(\mathbf{A}; \mathrm{L}_r(\mathbf{A})) > D_{KL}(\mathbf{A}; \mathbf{B})$.
Note that LrRR is a rank reduction method and does not perform matrix factorization. This means that we do not offer decomposed representation of $\mathrm{L}_r\left(\mathbf{A}\right)$ as a product of $n\times r$ and $r\times m$ matrices.

\section{Theoretical Analysis of L\protect\hidemath RR}\label{sec:4}

In this section, we derive our rank reduction algorithms for matrices
and show that L1RR is the best approximation in the sense of the KL divergence. In the following, we assume that an input to the algorithm is a positive matrix $\mathbf{A} \in \mathbb{R}_{> 0}^{n \times m}$. The rank reduction problem is to find a matrix $\mathbf{A}'\in \mathbb{R}^{n \times m}$ with the rank $r < \min\{n, m\}$ that approximates the input $\mathbf{A}$.

\subsection{Reminder on the Log-Linear Model on Posets}

We treat a matrix as a probability distribution whose sample space is matrix indices to use theoretical properties given in information geometry. To introduce our modeling, we overview the log-linear model on a poset~\citep{Sugiyama17ICML} in this subsection. This model is known to be a generalization of Boltzmann machines, where we can flexibly design interaction between variables using partial orders.
A poset $(S, \leq)$ is a set $S$ of elements associated with a partial order $\leq$ on $S$, where the relation ``$\leq$'' satisfies the following three properties: For all $x,y,z \in S$, (1) $x \leq x$, (2) $x \leq y, y\leq x \Rightarrow x = y$, and (3) $x \leq y,y \leq z \Rightarrow x \leq z$.
We consider a discrete probability distribution $p$ on a poset $(S,\leq)$, which is treated as a mapping $p:S \rightarrow (0,1)$ such that $ \sum_{x \in S} p(x) = 1$.
Each element $p(x)$ is assumed to be strictly larger than zero. We assume that the structured domain $S$ has the least element $\perp$; that is, $\perp \leq x$ for all $x \in S$.
The log-linear model for a distribution $p$ on $(S, \le)$ is defined as $\log{p(x)} = \sum_{s\leq x} \theta(s)$, where $\theta({\perp})$ corresponds to the normalizing factor (partition function). The convex quantity defined as the sign inverse $\psi(\theta) = -\theta(\perp)$ is called the Helmholtz free energy of $p$.

The canonical parameter $\theta$ of the log-linear model uniquely identifies the distribution $p$. Using $\theta$ as a coordinate system in the set of distributions, which is a typical approach in information geometry~\citep{Amari16}, we can draw the following geometric picture: Each point in the $\theta$-coordinate system corresponds to a distribution, and the resulting space is called the \emph{canonical space}.
Moreover, because the log-linear model belongs to the exponential family, we can also identify a distribution by expectation parameters defined as $\eta(x) = \sum_{x \leq s} p(s)$. Each expectation parameter $\eta(x)$ is literally consistent with the expected value $\mathbb{E}[F_x(s)]$ for the function $F_x(s)$ such that $F_x(s) = 1$ if $x \leq s$ and $0$ otherwise~\citep{Sugiyama16ISIT}. Thus we can also identify each point using the $\eta$-coordinate system in the \emph{expectation space}.
In addition, the $\theta$-coordinate and the $\eta$-coordinate are orthogonal with each other, which guarantees that we can combine these coordinates together as a mixture coordinate and a point specified by the mixture coordinate also identifies a distribution uniquely~\citep{Amari16}. As is clear from the definition, $\eta(\perp) = 1$ always holds.
In the following, we write a distribution as $p_\theta$ to emphasize that it is determined by the canonical parameter $\theta$. Similarly, we write $p_\eta$ if $p$ is determined by the expectation parameter $\eta$.

\subsection{Bingo Rule}\label{subsec:bingo_rule}

\begin{figure}[t]
\centering
\includegraphics[width=.9\linewidth]{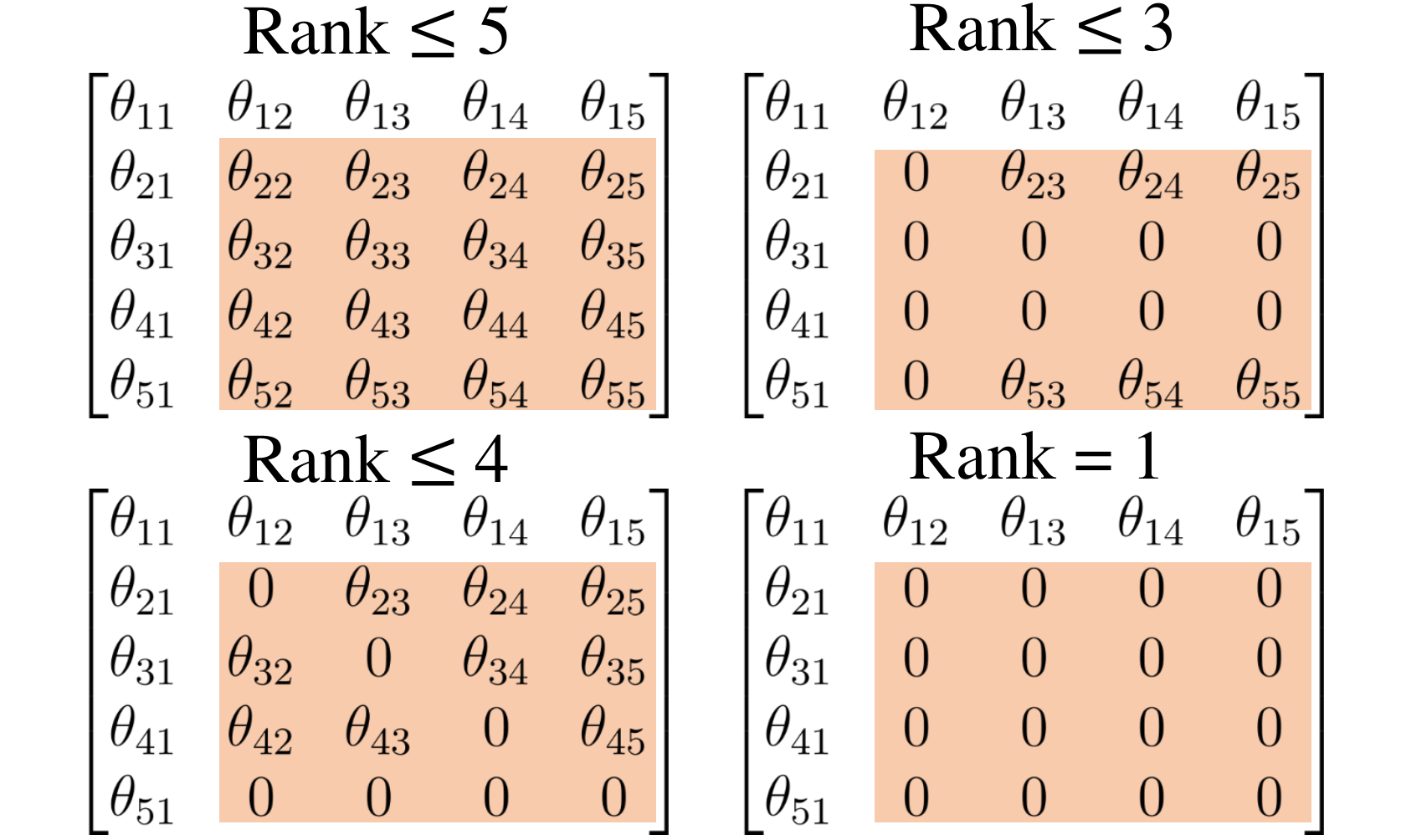}
\caption{The relationship between matrix ranks and the bingo rule in the case of $n=m=5$. Horizontal and vertical bingos reduce matrix rank. Bingos on the first column, the first row, and diagonal direction does not have any effect to the matrix rank. In the case of rank-$1$, the bingo selection is unique.}
\label{fig:bingo}
\end{figure}

To treat matrices by the log-linear model, we follow the approach proposed by~\citet{Sugiyama17ICML}.
Let us fix the domain $S = \{1, 2, \dots, n \} \times \{1, 2, \dots, m\} $, which is the set of indices of $ n \times m $ matrices.
We introduce a partial order as $(i,j) \leq (k, l)$ if $i \leq k$ and $j \leq l$ for all $(i, j), (k, l) \in S$, which allows us to make a mapping between a positive matrix $\mathbf{A}$ and its corresponding distribution $p$ as 
$p((i,j)) = \mathbf{A}_{ij}/\sum_{ij}\mathbf{A}_{ij}$. We write $p_{ij} = p((i,j))$, $\theta_{ij} = \theta((i,j))$, and $\eta_{ij} = \eta((i,j))$ to simplify notations.
The log-linear model for matrices is given as
\begin{align}\label{eq:loglinearMatrix}
    \log p_{ij} = \sum_{i' \le i} \sum_{j ' \le j} \theta_{i' j'}.
\end{align}
The value $\theta_{11} = \log p_{11}$ works as the normalizing factor of the model. It is known that this particular formulation can treat the problem of matrix balancing as optimization on the log-linear model~\citep{Sugiyama17ICML}. 
We can obtain the expectation parameters of the model as
\begin{align}\label{eq:loglinearEta}
    \eta_{ij} = \sum_{i \le i'} \sum_{j \le j'} p_{i' j'}.
\end{align}

By modeling matrices via the log-linear model on a poset, we can treat conditions on matrices through the pair of parameters $(\theta, \eta)$. Using this property, we introduce our key insight: the condition of low rank can be given as constraints on $\theta$, which we call the \emph{bingo rule}.

\begin{Definition.}[Bingo rule]
Given a positive matrix $\mathbf{A}\in\mathbb{R}^{n \times m}_{> 0}$ and its canonical parameter representation $\theta$. For $k\in\set{2, 3,\dots,m}$ and $j \in\set{2,3,\dots,n}$, we say that $k$-th column has a \emph{vertical bingo} if $\theta$ satisfies the condition 
\begin{align}\label{eq:bingo_k}
    \theta_{2k} = \theta_{3k} = \dots = \theta_{nk} = 0,
\end{align}
and $j$-th row has a \emph{horizontal bingo} if  
\begin{align}\label{eq:bingo_j}
    \theta_{j2}=\theta_{j3}=\dots=\theta_{jm}=0.
\end{align}
\end{Definition.} 

We do not consider bingos on the diagonal direction as they do not have direct connection to matrix ranks.
In addition, the bingo on the $1$-st column or the $1$-st row does not have any effect to the matrix rank.
We prove the relationship between vertical and horizontal bingos and ranks of matrices.

\begin{Theorem.}[Bingo rule and matrix rank]\label{theorem:bingo}
For a positive matrix $\mathbf{A}\in\mathbb{R}^{n \times m}_{> 0}$, when $\mathbf{A}$ has $v$ vertical bingos and $w$ horizontal bingos, it holds that
\begin{align*}
    \mathrm{rank}(\mathbf{A}) \leq \min\{n - w, m- v \}.
\end{align*}
\end{Theorem.}
\begin{proof}
 We show that, for every $j\in\set{2,3,\dots,n}$, a bingo on a $j$-th column implies that the $j$-th column of a given matrix is a multiplication of $(j-1)$-th column by a constant. From the definition of $p$, it holds that
 \begin{align}
    \frac{p_{i,j}}{p_{i,j-1}} = 
    \frac{\exp\left(\sum_{i'\leq i}\sum_{j'\leq j}\theta_{i'j'}\right)}
    {\exp\left(\sum_{i'\leq i}\sum_{j'\leq j-1}\theta_{i'j'}\right)} = \exp{\left(\sum_{i' \leq i} \theta_{i'j}\right)}.
    \label{eq:pijoverpij}
\end{align}
 If the $j$-th column has a vertical bingo, the right-hand side of Equation~\eqref{eq:pijoverpij} becomes $\exp(\sum_{i' \leq i} \theta_{i'j}) = \exp{(\theta_{1j})}$, which is a constant and does not depend on $i$, that is, the $j$-th column of a given matrix is multiplication of the $(j-1)$-th column by a constant $\exp{(\theta_{1j})}$. In the same way, for every $i\in\set{2,\dots,n}$, a horizontal bingo on an $i$-th row implies that the $i$-th row of given matrix is multiplication of the $(i-1)$-th row by a constant $\exp{(\theta_{i1})}$. Recalling the definition of the matrix rank, we have $\mathrm{rank}(\mathbf{A}) \leq \min\{n-w, m-v\}$ with $v$ vertical bingos and $w$ horizontal bingos.
\end{proof}

An example of the relationship between matrix ranks and  bingos is shown in Figure~\ref{fig:bingo}. Using the above theorem, we impose matrix constraints as $(\theta, \eta)$ conditions instead of directly imposing it on matrix elements, which enable us to apply information geometric techniques on matrix rank reduction. 

Although there are many possibilities to create rank-$r$ matrices using the bingo rule, it is in general hard to find the optimal bingo columns or rows as it costs $\Theta(2^r)$.
In our approach, we simply randomly sample bingo columns in \textbf{STEP1} in Section~\ref{sec:al} and empirically show that this heuristics still gives a good performance in Section~\ref{sec:exp}.

\subsection{Rank Reduction and Mean-Field Approximation}

We show that rank-$1$ approximation for matrices can be interpreted as a \emph{mean-field approximation}, which leads to the best rank-$1$ approximation in a closed form in Equation~\eqref{eq:l1rr_sol} via KL divergence minimization. Mean-field approximation approximates a given probability distribution with independent distributions. In the typical application to Boltzmann machines, which is defined as $p(\mathbf{x}) = \exp(\sum_ib_ix_i + \sum_{ij} w_{ij} x_i x_j)$ for a bias parameter $\mathbf{b} = (b)_i \in \mathbb{R}^{n}$, an interaction parameter $\mathbf{W} = (w_{ij})\in \mathbb{R}^{n \times n}$, and a binary vector $\mathbf{x} \in \set{0, 1}^n$, the mean-field equation is given as $\eta_i = \sigma(b_i + \sum_j w_{ij} \eta_j)$. To approximate a Boltzmann machine $p$ by independent distributions $p_1, p_2, \dots, p_n$, we need the expected value $\mathbb{E}_p[x_i]$ for each $i \in \{1, \dots, n\}$, each of which requires exponential computational cost $O(2^n)$~\citep{anderson1987mean}. To avoid this expensive computational cost, the mean-field equation is numerically solved to obtain the approximation value of the expected value $\mathbb{E}_p[x_i]$. In contrast, in our modeling, we will show in Theorem~\ref{th:L1RR} that the expected values correspond to the row-sums and column-sums of a matrix, hence their computation takes only $O(nm)$. Thus \emph{we can conduct mean-field approximation without solving the mean field equation}, which makes our rank-$1$ reduction faster. 

Theoretical properties of mean-field approximation have been analyzed in information geometry, where it can be understood as a projection onto a submanifold described by the pair $(\theta, \eta)$ of canonical and expectation parameters~\citep{tanakaTheoryMeanField1999}. This submanifold has a special property that $\eta$ can be easily computed from $\theta$ as discussed for the typical Boltzmann machines. We point out the analogy that rank-$1$ reduction can also be captured as projection onto a submanifold in which we can obviously know expectation parameters $\eta$ from the canonical parameters $\theta$.

To achieve rank reduction in the submanifold of canonical parameters $\theta$, first we prepare the set of distributions that corresponds to the set of rank-$r$ matrices using the Bingo rule.
From Theorem~\ref{theorem:bingo} and the conditions in Equations~\eqref{eq:bingo_k} and~\eqref{eq:bingo_j}, let us define the \emph{model submanifold}
\begin{align*}
    \mathcal{P}_r = \{ p_\theta \mid p \text{ has } m-r \text{ vertical or horizontal bingos}\},
\end{align*}
which is obtained by linear constraints on $\theta$ and therefore is convex with respect to $\theta$.
From Theorem~\ref{theorem:bingo}, for any distribution in $\mathcal{P}_r$, it is guaranteed that the rank of the corresponding matrix is less than or equal to $r$.
In contrast, the $\eta$-coordinate allows us to formulate another submanifold from a given matrix $\mathbf{A}$.
Let $\eta^\mathbf{A}$ be the $\eta$ representation of $\mathbf{A}$. The \emph{data submanifold} is defined as 
\begin{align*}
    \mathcal{P}_\mathbf{A} = \Set{ p_\eta |
    \begin{aligned}
    & \eta_{ij} = \eta_{ij}^\mathbf{A} \text{ if } i = 1, j = 1,\\
    &\text{or } j \text{ is non-bingo column index}
    \end{aligned}
    }.
\end{align*}
In our formulation, rank reduction of a given matrix is understood to be a projection of some initial distribution in the model submanifold $\mathcal{P}_r$ onto the data submanifold $\mathcal{P}_{\mathbf{A}}$.
This operation is known as \emph{$e$-projection} in information geometry, which is a convex optimization and ensures that the global optimum is always closest to the input matrix in the sense of the KL divergence.
Gradient based methods such as natural gradient can be used to achieve $e$-projection of some distribution in the specified bingo subspace $\mathcal{P}_r$ onto the data submanifold $\mathcal{P}_{\mathbf{A}}$.
However, the number of parameters $\theta_{ij}$ we have to optimize is $nr+m-r$, and the natural gradient has $O((nr+m-r)^3)$ cost for each iteration. In fact, Legendre decomposition~\citep{sugiyama2019legendre} also conducts the $e$-projection, and its computation is not efficient as we will show in Section~\ref{sec:exp}.

To avoid such expensive computation in $e$-projection, we solve the rank-$1$ approximation via a mean-field approximation.
In the following, we first discuss the case of rank-$1$ reduction. Our strategy is to employ $m$-projection instead of the typical $e$-projection. The $m$-projection is a backward direction of $e$-projection, that is, it is 
a projection from a given distribution $\mathbf{A} \in \mathcal{P}_\mathbf{A}$ to the bingo submanifold $\mathcal{P}_1$ and it also minimizes the KL divergence. By using the decomposability of probability and its expectation parameters $\eta$, we can obtain the destination of the $m$-projection onto $\mathcal{P}_1$ in a closed form. As a result, we obtain an \emph{analytical solution} of the projection destination.
After establishing the rank-1 reduction, we develop the general case of rank-$r$ reduction by multiple application of rank-$1$ reduction.


Let us consider the rank-1 submanifold $\mathcal{P}_1$. We also call $\mathcal{P}_1$  \emph{full bingo space}. 
From its definition, the canonical parameter $\overline\theta$ of a distribution $\overline p = p_{\overline\theta} \in \mathcal{P}_1$ always satisfies the condition $\overline\theta_{ij} = 0$ if $i \neq 1$ and $j \neq 1$. Therefore each probability can be directly expressed as $\overline p_{ij} = \exp({\overline\theta_{11}})\exp{(\sum_{i'=2}^i\overline\theta_{i'1}+\sum_{j'=2}^j\overline\theta_{1j'}})$.
Using this property, we can decompose any distribution $\overline{p} \in \mathcal{P}_1$ into the product of the following two independent probability vectors $\overline{\mathbf{p}}^{(1)} \in\mathbb{R}^n$ and $\overline{\mathbf{p}}^{(2)} \in\mathbb{R}^m$,
whose elements are given as\footnote{Note that the empty sum is treated as zero, $\sum_{x \in \emptyset} f(x) = 0$ for any mapping $f(\cdot)$ and the empty set $\emptyset$.}:
\begin{align}
    \overline p^{(1)}_{i} &= 
    \frac{ \exp{( \sum_{k=2}^i\overline\theta_{k1})}}
    { 1 + \sum_{k=2}^n \exp(\sum_{m=2}^k \overline\theta_{m1})}, \label{eq:ind_j}\\
    \overline p^{(2)}_{j} &= 
    \frac{ \exp{( \sum_{k=2}^j \overline\theta_{1k})}}
    { 1 + \sum_{k=2}^m \exp(\sum_{n=2}^k \overline\theta_{n1})}. \label{eq:ind_i}
\end{align}
Each element in $\overline{\mathbf{p}}^{(1)}$ depends on only $\overline\theta_{21},\overline\theta_{31},\dots,\overline\theta_{n1}$ and
each element in $\overline{\mathbf{p}}^{(2)}$ depends on only $\overline\theta_{12},\overline\theta_{13},\dots,\overline\theta_{1m}$. We can also regard these two vectors as discrete probability distributions since they are normalized, that is, $\sum_i\overline{p}_i^{(1)}=\sum_j\overline{p}_j^{(2)}=1$.
As a result, the following proposition holds.

\begin{Proposition.}  
For any distribution $\overline{p} = p_{\overline\theta} \in \mathcal{P}_1$, if we treat it as an $n \times m$ rank-$1$ matrix, it can be decomposed as $\overline{p} = \overline{\mathbf{p}}^{(1)}\overline{\mathbf{p}}^{(2) \mathrm{T}}$.
\end{Proposition.}  

Using the above proposition, we get the rank-$1$ condition as an $\eta$ expression.
\begin{Theorem.}\label{th:expect}
For every distribution $\overline p \in \mathcal{P}_1$, its expectation parameter $\overline \eta$ can be decomposed as $\overline\eta_{ij} = \overline\eta_{i1}\overline\eta_{1j}$. 
\end{Theorem.}
\begin{proof}
We show the decomposability of the distributions on $\mathcal{P}_1$ into $\overline{p}_{ij} = \overline{p}^{(1)}_i \overline{p}^{(2)}_j$. Using the independence of probability and the normalization condition $\sum_{1 \leq i'}^n\sum_{1 \leq j'}^n \overline{p}_{i'j'}=1$, it follows that 
\begin{align*}
\overline\eta_{ij} 
&= \sum_{i \leq i'}\sum_{j \leq j'} \overline{p}_{i'j'} 
= \left(\sum_{i \leq i'} \overline{p}^{(1)}_{i'}\right) \left(\sum_{j \leq j'} \overline{p}^{(2)}_{j'} \right) \\
&= \left(\sum_{i \leq i'} \overline{p}^{(1)}_{i'}\right) \left(\sum_{1 \leq i'} \overline{p}^{(1)}_{i'} \sum_{1 \leq j'}\overline{p}^{(2)}_{j'}\right) \left(\sum_{j \leq j'} \overline{p}^{(2)}_{j'}\right) \\
&= \left(\sum_{i \leq i'} \overline{p}^{(1)}_{i'} \sum_{1 \leq j'} \overline{p}^{(2)}_{j'} \right)\left(\sum_{1 \leq i'} \overline{p}^{(1)}_{i'} \sum_{j \leq j'} \overline{p}^{(2)}_{j'}\right) 
= \overline\eta_{i1}\overline\eta_{1j}.
\end{align*}
Therefore $\overline\eta_{ij} = \overline\eta_{i1}\overline\eta_{1j}$ holds.
\end{proof}

Our result means that, if a matrix is reduced to rank-$1$, it can be always decomposed 
into the product of independent distributions. Hence rank-$1$ reduction can be viewed as mean-field approximation.
This is consistent with the fundamental property known in linear algebra such that any rank-$1$ matrix can be expressed as Kronecker product of two vectors, which correspond to (\ref{eq:ind_j}) and (\ref{eq:ind_i}) in our case. 

Our formulation of matrices via the log-linear model on a poset enables us to find the optimal solution via $m$-projection without solving the mean field equation.
It has been shown that the expectation values of each random variable does not change before and after the $m$-projection~\citep{Amari16}. That is, in the $m$-projection from $\eta$ to the full bingo space $\mathcal{P}_1$ in our log-linear model for a positive matrix, $\eta_{i1} = \overline\eta_{i1}$ and $\eta_{1j} = \overline\eta_{1j}$ hold. Here,$\overline\eta$ is the projection destination of the $m$-projection. Using this property of $m$-projection, we derive L$1$RR as follows.

\begin{figure}[t]
\centering
\includegraphics[width=.9\linewidth]{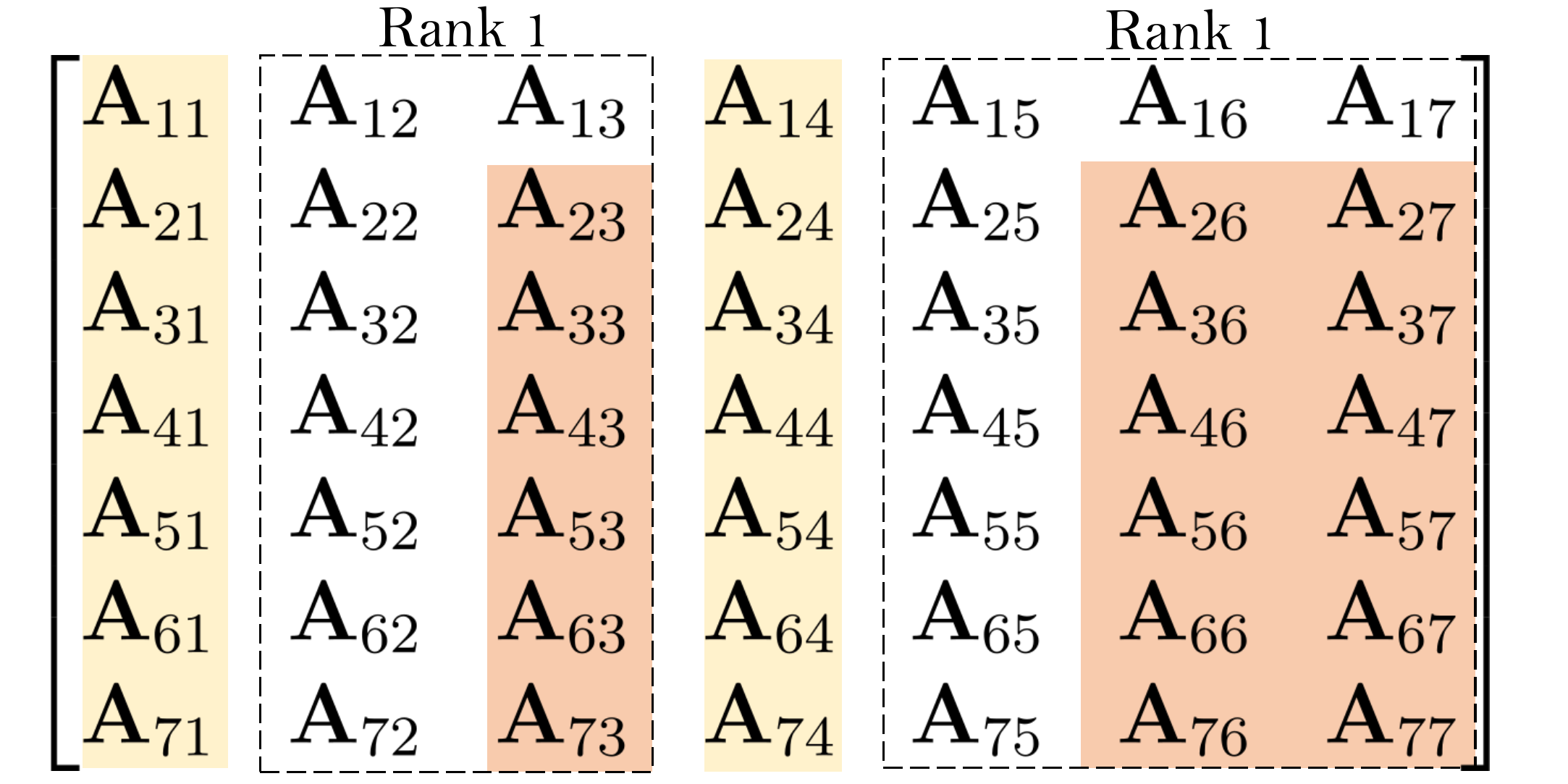}
\caption{Example of L4RR in the case of $n=m=7$ and $C=\set{3,6,7}$. Orange areas has $\theta$ bingos and yellow areas preserve $\eta_{ij}$. By imposing three bingos on input $\mathbf{A}$, the matrix rank is reduced by three at most.}
\label{fig:sketch}
\end{figure}

\begin{Theorem.}\label{th:L1RR}
When $\overline p \in \mathcal{P}_1$ is the destination of the $m$-projection onto full bingo subspace $\mathcal{P}_1$ from a distribution $p$, $\overline p$ is given as follows:
\begin{align}\label{eq:L1RR}
    \overline{p}_{ij} = \biggl( \sum_{i'} p_{i'j} \biggl) \biggl( \sum_{j'} p_{ij'} \biggr).
\end{align}
\end{Theorem.}  
\begin{proof}
    During the $m$-projection, $\eta_{i1}$ and $\eta_{1j}$ for ${i,j} \in \set{2,\dots, n} \times \set{2,\dots,m}$ does not change. By using the rank-$1$ condition $\overline\eta_{ij}=\overline\eta_{i1}\overline\eta_{1j}$, it follows that
    \begin{align}\label{eq:th6}
        \overline p_{ij} & = 
         \overline \eta_{ij}  
         - \overline \eta_{i,j+1}  
         - \overline \eta_{i+1,j}  
         + \overline \eta_{i+1,j+1}  
        \nonumber\\
        &=
         \left( \eta_{i, 1} -  \eta_{i+1, 1}  \right)
         \left( \eta_{1, j} -  \eta_{1, j+1}  \right)
         \nonumber\\
        &=
         \Bigl( \sum_{i'} p_{i'j} \Bigl) \Bigl( \sum_{j'} p_{ij'} \Bigl).
    \end{align}
    The first line is derived from the definition of the expected parameters, 
    $\eta_{ij}=\sum_{i \leq i',j \leq j'}p_{i'j'}$.
\end{proof}

Theorem~\ref{th:L1RR} guarantees L$1$RR introduced in Section~\ref{sec:21}. If a given matrix is not normalized, we need to divide the right-hand side of Equation~\eqref{eq:L1RR} by the sum of all entries of the matrix in order to match the scales of input and output. In the case of rank-$1$ reduction, the possible bingo columns and rows are unique; that is, $\theta_{ij} = 0$ if $i > 1$ and $j > 1$ (see the right lower matrix in Figure~\ref{fig:bingo}). Because $m$-projection finds the global minimum, it is guaranteed that the approximation minimizes the KL divergence from an input distribution (normalized matrix) $p$; that is, 
\begin{align*}
\overline{p} = \argmin_{\overline{p}' \in \mathcal{P}_1} D_{KL} ( p ; \overline{p}') \iff \overline{p} = \Bigl( \sum_{i'} p_{i'j} \Bigl) \Bigl( \sum_{j'} p_{ij'} \Bigl).
\end{align*}

Next we extend the above discussion to rank-$r$ reduction. We formulate rank-$r$ reduction of an input matrix $\mathbf{A}$ as a $m$-projection to the $m-r$ bingos submonifold $\mathcal{P}_r$. To simplify the following discussion, we assume that the model submanifold $\mathcal{P}_r'$ is given as
\begin{align}\label{eq:bingo_fix_right}
    \mathcal{P}_r'=\{\, p_\theta \mid \theta_{2j}=\theta_{3j}=\dots=\theta_{nj}=0 \text{ if } j > r \,\},
\end{align}
where $\mathcal{P}_r' \subseteq \mathcal{P}_r$.
We show that the $m$-projection onto $\mathcal{P}_r'$ can be archived by $\mathrm{L}_1(\mathbf{A})$ on a submatrix of $\mathbf{A}$. Since the $m$-projection does not change the part of $\eta$ parameters $\{ \eta_{ij} \mid i = 1 \text{ or } j \leq r \}$ and the relation between each element of $\mathbf{A}$ and its $\eta$-parameters is described as $p_{ij}=\eta_{ij}-\eta_{i,j+1}-\eta_{i+1,j}+\eta_{i+1,j+1}$ for $p_{ij}=\mathbf{A}_{ij}/\sum_{ij}\mathbf{A}_{ij}$, the part of the elements of an input matrix $\{ \mathbf{A}_{ij} \mid j < r \}$ does not change in the $m$-projection. Then, we can reduce the matrix rank of $\mathbf{A}$ to $r$ with keeping the values of $\mathbf{A}_{1:r-1}$, where the matrix rank of $\mathbf{A}_{r:m}$ is reduced to 1. Therefore we conduct L1RR on the submatrix $\mathbf{A}_{r:m}$ and replace $\mathbf{A}_{r:m}$ with $\mathrm{L}_1(\mathbf{A}_{r:m})$ to obtain rank reduced matrix. The replacement corresponds to \textbf{STEP3} in Section~\ref{sec:al}. If the bingo position is not fixed in one place like Equation~\eqref{eq:bingo_fix_right}, we perform the same procedure for each piece. Figure~\ref{fig:sketch} is the sketch of the LrRR in such a case.

The above discussion guarantees that LrRR achieves the global minimum on the selected bingos model space $\mathcal{P}_r$ since the $m$-projection finds the minimum point of the KL divergence. As we have already shown, L1RR is always the best rank-1 positive approximation in the sense of KL divergence as the bingo selection is unique when $r = 1$.
We can say that rank-$r$ reduction is multiple partial mean-field approximations for submatrices as we formulate LrRR by combining rank-$1$ reduction.

Since rank reduction is formulated as $m$-projection that keeps a part of $\eta$, interestingly, Legendre rank-$r$ reduction preserves the row sums and the column sums, that is,
\begin{align*}
\sum_i {\rm L}_r(\mathbf{A})_{ij} = \sum_i \mathbf{A}_{ij},\quad
\sum_j {\rm L}_r(\mathbf{A})_{ij} = \sum_j \mathbf{A}_{ij}.
\end{align*}
In fact, it has already been shown that row sums and column sums are conserved in KL-NMF~\citep{ho2008non}. We naturally rederived the same consequence from the geometric viewpoint using the property of $m$-projection. Furthermore, we can confirm that $\mathrm{L}_1(\mathbf{A})$ is a constant multiple of an integer matrix when the input matrix $\mathbf{A}$ is an integer matrix. The rank reduction of integer matrices is still uncharted territory and is interesting future work.

The idea of matrix and tensor decomposition in the framework of information geometry has already been discussed in ~\citep{sugiyama2019legendre}. However, we firstly find the bingo rule, and derived a method to achieve the projection to bingo space only by computing the row sum and the column sum of the input matrix without using an iterative method such as gradient descent. As experimental results in Section~\ref{sec:exp} show, our method is much faster and accurate than Legendre decomposition~\cite{sugiyama2019legendre}.

\section{Numerical Experiments}\label{sec:exp}

\begin{figure*}[t]
\centering
\includegraphics[width=.99\linewidth]{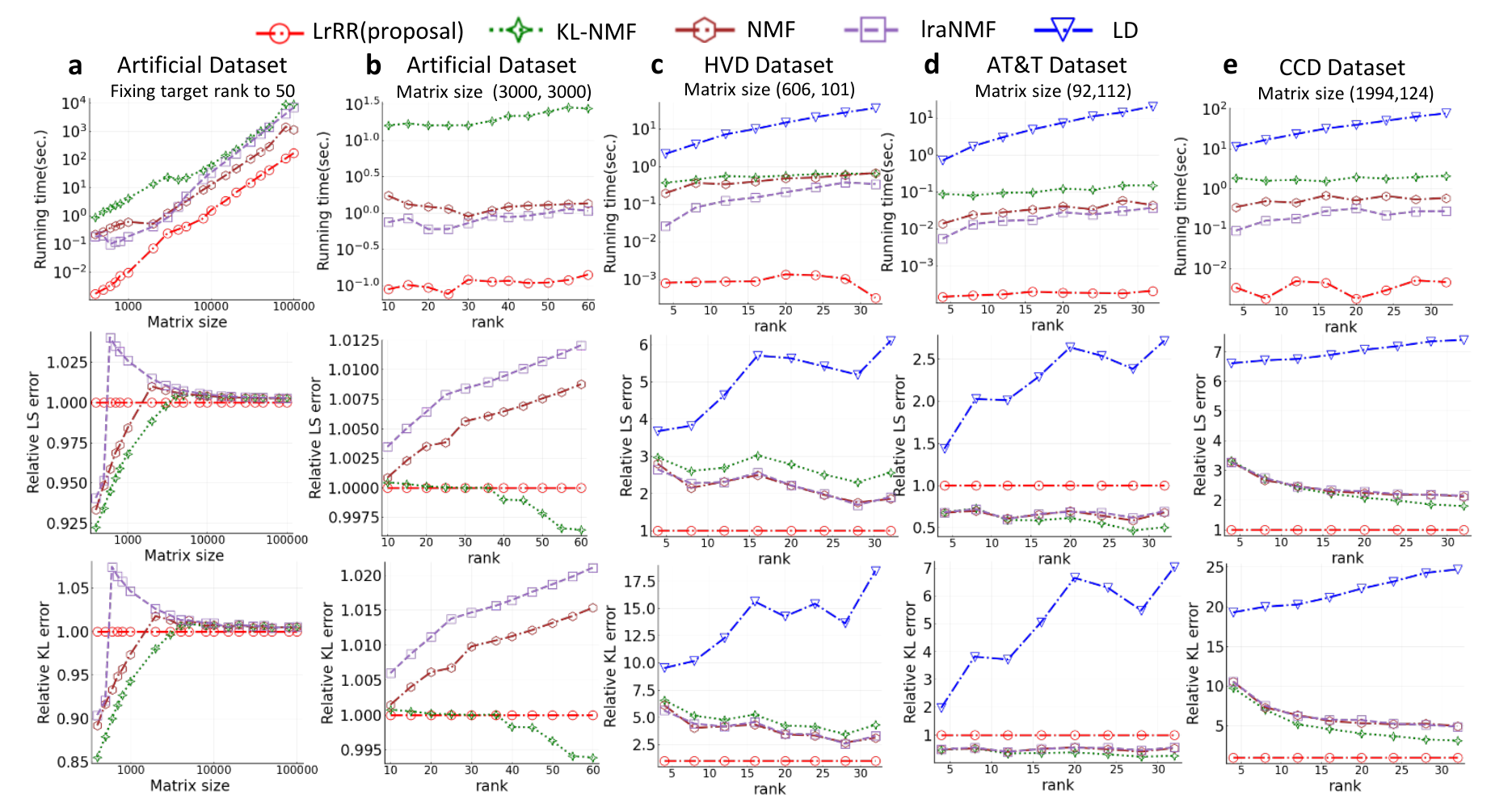}
\caption{Experimental results. These results show that our proposal LrRR (red circle) is consistently faster than other methods with competitive or better LS (least square) and KL divergence errors defined in Equations~\eqref{eq:errorscore_KL} and~\eqref{eq:errorscore_LS}.}
\label{fig:expresult}
\end{figure*}

We empirically evaluate the performance of LrRR using synthetic and real-world datasets. As comparison partners, we use NMF~\citep{paatero1994positive} and lraNMF~\citep{zhou2012fast}. lraNMF is known as a faster approximation method of NMF. Since the cost function of the above two methods is the Least Squares (LS) function, we also compare KL-NMF~\citep{dhillon2005generalized,ho2008non} and Legendre decomposition (LD)~\cite{sugiyama2019legendre} for matrices, which use the KL divergence as the cost function.

To measure the quality of the obtained low rank matrices, we use the relative LS error $\phi_{\rm LS}(\mathbf{A})$ between an input matrix and an obtained rank reduced matrix and the relative KL divergence $\phi_{\rm KL}(\mathbf{A})$ from an input matrix to an obtained rank reduced matrix. More precisely, we define these scores as
  \begin{align}\label{eq:errorscore_KL}
         \phi^{\rm method}_{\rm KL}(\mathbf{A})
         &= 
         \frac
         {D_{\rm KL}\left(\mathbf{A} ; \mathbf{A}^{\rm method}\right)}
         {D_{\rm KL}\left(\mathbf{A} ; {\rm L}_r(\mathbf{A})\right)}, \\
         \label{eq:errorscore_LS}
        \phi^{\rm method}_{\rm LS}(\mathbf{A})
        &= 
         \frac
         {\| \mathbf{A} - \mathbf{A}^{\rm method} \|_{\mathrm F}}
         {\| \mathbf{A} - {\rm L}_r(\mathbf{A}) \|_{\mathrm F}}.
    \end{align}
The score 1.0 means that the method has the same performance with LrRR, and larger values than 1.0 means that our method is better.

All of these methods are implemented in Julia 1.5. Experiments were conducted on CentOS 6.10 with a single core of 2.2 GHz Intel Xeon CPU E7-8880 v4 and 3TB of memory. We implement LD by directly translating the original \texttt{C++} code provided in~\citep{sugiyama2019legendre} into \texttt{Julia}. We implement lraNMF by referring to the pseudo code in the original paper~\citep{zhou2012fast}. We implement KL-NMF and NMF by directly translating the \texttt{sklearn} code ~\citep{scikit-learn} into \texttt{Julia}. 

\paragraph{Experiments on Synthetic Data}
First we examine the scalability of our method with varying the input matrix size $n$ for $n\times n$ matrices with keeping the target matrix rank $r=50$. An input matrix is generated by the multidimensional uniform distribution from 0.0 to 1.0. Results are illustrated in Figure~\ref{fig:expresult}(a). As we can see on the top panel of Figure~\ref{fig:expresult}(a), in terms of the efficiency, LrRR is much faster than comparison partners. This is because our method is not an iterative method unlike other methods. Middle and bottom plots show that the KL and LS errors of other methods asymptotically approach the error of LrRR as the size of the matrix increases. 

Next, we examine the efficiency and the effectiveness of rank reduction methods with varying the target rank $r$ with keeping the matrix size $n=3,000$ of synthetic matrices generated by the multidimensional uniform distribution from 0.0 to 1.0. Results are illustrated in Figure~\ref{fig:expresult}(b). KL-NMF tries to find a rank-$r$ matrix with considering any rank-$r$ matrices, while LrRR achieves rank-$r$ approximation by a combination of rank-$1$ approximations. Therefore, it is reasonable that KL-NMF can reach a better approximation. Still, LrRR is competitive with KL-NMF when tha target rank is small and better than other methods. Note that, in these two experiments, matrices size is too large to adapt LD. This is why we did not conduct experiments with LD. 

\paragraph{Experiments on Real Data}
We examine the behaviour of the proposed method on real world datasets. We use 
Hill-Valley Data Set (HVD)~\citep{Dua:2019}, Database of Faces (AT\&T)~\citep{samaria1994parameterisation}, and Communities and Crime Open Data Set (CCD)~\citep{united19921990}. We use the entire CCD to make a $1994 \times 124$ matrix. AT\&T includes 400 grey-scale face photos. The size of each image is $(92,112)$. We use the first image \texttt{s1/s1.pgm} to make a $92 \times 112$ matrix. We use \texttt{Hill\_Valley\_with\_noise\_Testing\.data} in HVD to make a $606\times 101$ matrix. 

In all datasets, LrRR is always an order of magnitude faster than comparison partners. In terms of the effectiveness of LrRR, not only are the KL errors competitive but also the LS error relative to other methods. When the ratio of the target rank to the matrix size is small, LrRR tends to have better approximation performance than other methods.
In addition, three matrices are small enough to apply LD. However, experimental results in Figure~\ref{fig:expresult}($\boldsymbol{\rm c,d,e}$) show that, even when the target rank $r$ is small, LD is much slower and less accurate than our proposed method.

\section{Connection between Rank Reduction and Matrix Balancing}

In this section, we further investigate geometric property of Legendre rank reduction.
In particular, we uncover the relationship between rank-$1$ reduction and \emph{matrix balancing}, which apparently seem to be unrelated linear algebraic tasks.

First we introduce the $(\mathbf{s},\mathbf{t})$-balancing problem on a given normalized positive matrix $\mathbf{B} \in \mathbb{R}_{> 0}^{n \times n}$ for $\mathbf{s}, \mathbf{t} \in \mathbb{R}^{n}_{> 0}$. The problem of $(\mathbf{s}, \mathbf{t})$-balancing of a matrix $\mathbf{B}$ is defined as that of finding $\mathbf{B}' = {\rm diag}(\mathbf{u}) \mathbf{B} {\rm diag}(\mathbf{v})$ for $\mathbf{u}, \mathbf{v} \in \mathbb{R}^{n}$ such that  $\sum_i \mathbf{B}'_{ij} = s_j$ and $\sum_i \mathbf{B}'_{ij} = t_j$. Note that there is no solution when $\sum_i s_i \neq \sum_j t_j$, hence we always assume that $\sum_i s_i = \sum_j t_j = 1$ without loss of generality. As well as rank reduction represented by constraints on $\theta$, the balancing condition is given as constraints on $\eta$ as
\begin{align}\label{eq:eta_balancing}
\eta_{i1} = \sum_{i'=i}^n s_{i'} , \quad
\eta_{1j} = \sum_{j'=j}^m t_{j'},
\end{align}
yielding the \emph{balancing submanifold} $\mathcal{P}_B = \{ p(\eta) \mid \eta$ satisfies the condition~\eqref{eq:eta_balancing} $\}$.

By considering balancing and rank reduction simultaneously in the framework of information geometry, we can derive the following property that the balanced rank-$1$ matrix always uniquely exists.

\begin{Theorem.}\label{th:t5}
    The intersection $\mathcal{P}_B \cap \mathcal{P}_1$ is a singleton.
\end{Theorem.} 

\begin{proof}
    As we discussed in Section~\ref{sec:4}, we can identify a distribution using the mixture coordinate system $(\theta, \eta)$ that combines $\theta$- and $\eta$-coordinates. Therefore, specifying $n \times m$ parameters on the mixture coordinate $(\theta, \eta)$ uniquely identifies a matrix $\mathbf{A} \in \mathbb{R}^{n \times m}$. We can see that balancing condition in Equation~\eqref{eq:eta_balancing} determines $n+m-1$ parameters; that is, $\eta_{i1}$ and $\eta_{1j}$ for $i \in \set{2,\dots,n}$, $j \in \set{2,\dots,m}$, and rank-$1$ condition determines the rest $(n-1)\times(m-1)$ parameters; that is, $\theta_{ij} = 0$ for $(i,j) \in \set{2,\dots,n} \times \set{2,\dots,m}$. The normalizing factor determines $\theta_{11}$ and $\eta_{11}$. Now, the balancing conditions and the rank-$1$ condition specify all $n \times m$ parameters, therefore the mixture coordinate $(\theta,\eta)$ uniquely identifies the rank-$1$ balanced matrix.
\end{proof}

Moreover, we have an analytical solution to the matrix $\{\mathbf{A}^{\prime}\} = \mathcal{P}_B \cap \mathcal{P}_1$, which is obtained as $\mathbf{A}^{\prime}_{ij} = s_it_j$.


\begin{figure}[t]
\centering
\includegraphics[width=.99\linewidth]{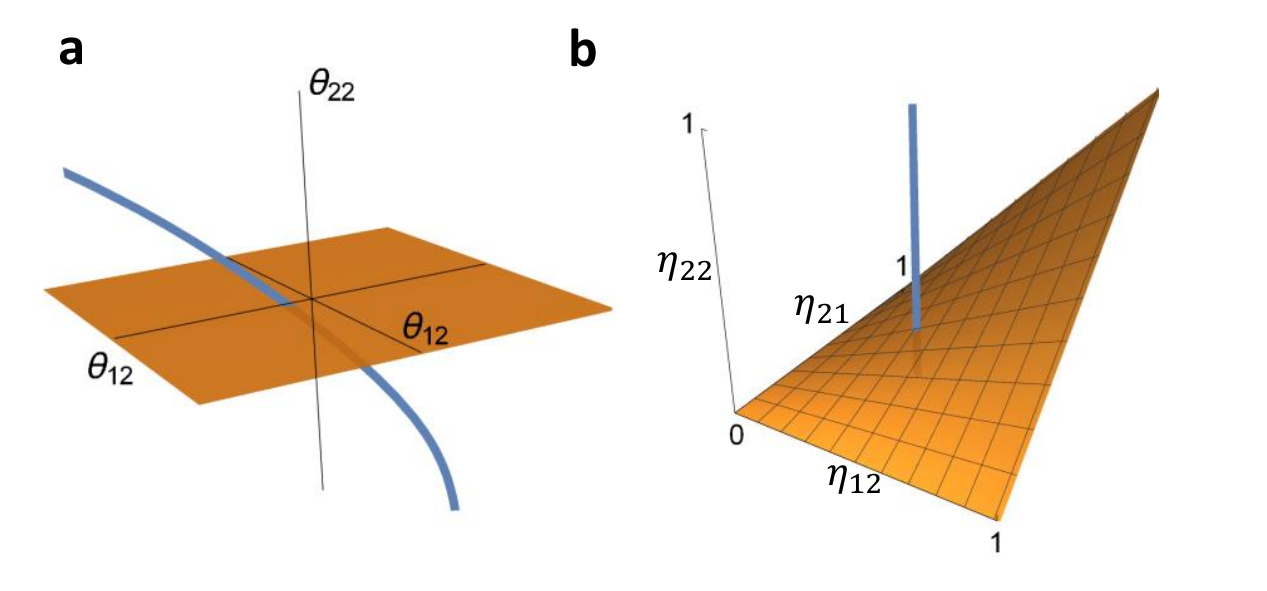}
\caption{Balancing submanifold $\mathcal{P}_B$ (blue) and rank-$1$ submanifold (orange) $\mathcal{P}_1$ in $\theta$ space (a) and $\eta$ space (b) with $n=m=2$ and $\mathbf{s}=\mathbf{t}=(0.4,0.6)$.}
\label{fig:crosspoint}
\end{figure}


To get the intuition of geometric structure across conditions on rank-$1$ reduction, matrix balancing, and mean-field approximation, we illustrate a simple case of $n = 2$ as 3D plots in Figure~\ref{fig:crosspoint}. Let us consider the $(\mathbf{s}, \mathbf{t})$-balanced matrix $\mathbf{B} \in \mathbb{R}_{> 0}^{2\times2}$ with $n = 2$. From the balancing condition, we can get the following analytical solution using $\mathbf{B}_{22}$:
\begin{align*}
    \mathbf{B} &= 
    \begin{bmatrix}
    1-s_2-c_2+\mathbf{B}_{22} & t_2 - \mathbf{B}_{11}\\
        s_2 - \mathbf{B}_{22} & \mathbf{B}_{11} \\
    \end{bmatrix} ,\\
    \theta^{\mathbf{B}} &=
    \begin{bmatrix}
        \log{(1-s_2-t_2+\mathbf{B}_{22})}     & \log{\frac{t_2-\mathbf{B}_{22}}{(1-s_2-t_2+\mathbf{B}_{22})}}  \\
        \log{\frac{t_2-\mathbf{B}_{22}}{(1-s_2-t_2+\mathbf{B}_{22})}}  & \log{\frac{\mathbf{B}_{22}(1-s_2-t_2+\mathbf{B}_{22})}{(t_2-\mathbf{B}_{22})(s_2-\mathbf{B}_{22})}}  \\
    \end{bmatrix}
    , \\ 
    \eta^{\mathbf{B}} &=
    \begin{bmatrix}
        1     & t_2  \\
        s_2  & \mathbf{B}_{22}  \\
    \end{bmatrix}.
\end{align*}
Remember that $\theta_{11}$ corresponds to the normalizing factor and $\eta_{11} = 1$. The submanifold consisting of balanced matrices can be drawn as a convex curve in a 3-dimensional space by regarding $\mathbf{B}_{11}$ as a mediator variable. Interestingly, the curve becomes a straight line in the $\theta$ space only when $\mathbf{s} = \mathbf{t} = (0.5,0.5)$. In contrast, the set of rank-$1$ matrices is identified as a plane $(\theta_{21},\theta_{12},0)$ in the $\theta$ space since $\theta_{22}=0$ ensures ${\rm rank}(\mathbf{A})=1$ and on the plane $(\eta_{21},\eta_{12},\eta_{21}\eta_{12})$ in the $\eta$ space (Theorem~\ref{th:expect}). We observe that balanced space $\mathcal{P}_B$ and mean-field space $\mathcal{P}_1$ cross a point, which is shown in Figure~\ref{fig:crosspoint}. It is coherent with Theorem~\ref{th:t5}. The cross point dynamically changes by $\mathbf{s}$, $\mathbf{t}$.

\section{Conclusion}
In this paper, we have proposed a new rank reduction method for matrices, called Legendre rank reduction. Our key idea is to realize the low rank condition on not matrices directly but the canonical parameter space of the log-linear model on a poset, where each matrix is treated as a discrete probability distribution. Our theoretical contribution is that we have firstly shown the direct relationship between three different problems: rank reduction of matrices, mean-field approximation, and matrix balancing, using information geometry. This connection enables us to formulate an efficient rank reduction algorithm, which analytically computes the optimal solution that minimizes the KL divergence from a given matrix. Our geometric analysis uncovered the following statement: rank-$1$ reduction coincides with mean-field approximation. Furthermore, we have empirically shown the efficiency of our rank reduction method compared to NMF-based methods. Our work will become a basis of further investigation between linear algebraic matrix operations, statistics, and machine learning via information geometry.





\bibliography{main}

\appendix
\providecommand{\upGamma}{\Gamma}
\providecommand{\uppi}{\pi}

\end{document}